\documentclass{cys}

\usepackage[utf8]{inputenc}

\usepackage[russian,english]{babel}  

\usepackage{url}
\usepackage[noabbrev,capitalise]{cleveref}
\usepackage{graphicx}
\usepackage{multirow}
\usepackage{xcolor}
\usepackage{pgfplots}
\pgfplotsset{compat=1.14}
\usepackage{xspace}
\usepackage{tikz-dependency}
\usepackage{caption}
\usepackage{subcaption}

\usepackage{epstopdf} 

\addto\captionsenglish{%
}

\addto\captionsspanish{%
}

\title{
Promoting the Knowledge of Source Syntax in Transformer NMT
\\ Is Not Needed
}

\author{Thuong-Hai Pham, Dominik Mach\'{a}\v{c}ek, Ond\v{r}ej Bojar}

\affil{ 
  Charles University \authorcr
   Faculty of Mathematics and Physics \authorcr
Institute of Formal and Applied Linguistics \authorcr
Prague, Czech Republic

\authorcr  \authorcr
\{pham,machacek,bojar\}@ufal.mff.cuni.cz
\authorcr  \authorcr
}

\begin{document}

\def\XXX#1{\textcolor{red}{#1}}
\def\hideXXX#1{}
\def\fXXX#1{\XXX{\footnote{#1}}}
\def\furl#1{\footnote{\url{#1}}}
\def\parcite#1{\cite{#1}} 
\def\perscite#1{\cite{#1}} 
\def\inparcite#1{\cite{#1}} 

\def\dummy#1{``{#1}''}
\def\de2cs{de2cs\xspace}
\def\cs2en{cs2en\xspace}
\def\transformer{Transformer\xspace}
\def\transformerbase{TransformerBase\xspace}
\def\transformerrel{\texttt{TransformerRelative}\xspace}
\def\seq2seq{\texttt{seq2seq}\xspace}
\def\DiagonalParse{DiagonalParse\xspace}
\def\DepParse{DepParse\xspace}

\def\citet#1{\cite{#1}}
\def\citep#1{\cite{#1}}
\def\Sref#1{Section~\ref{#1}}
\def\Tref#1{Table~\ref{#1}}
\def\Fref#1{Figure~\ref{#1}}

\maketitle

\renewcommand{\tablename}{Table}

\begin{abstract}
The utility of linguistic annotation in neural machine translation seemed to
had been
established in past papers. The experiments were however limited to recurrent
sequence-to-sequence architectures and relatively small data settings.

We
focus on the state-of-the-art Transformer model and use comparably larger
corpora. Specifically, we try to promote the knowledge of source-side
syntax using multi-task learning either through simple data manipulation
techniques or through a dedicated model component. In particular, we train one
of Transformer attention heads to produce source-side dependency tree.

Overall, our results cast some doubt on the utility of multi-task setups with
linguistic information. The data manipulation
techniques, recommended in previous works, prove 
ineffective in large data settings.

The
treatment of self-attention as dependencies seems much more promising: it helps in translation and reveals
that Transformer model can very easily grasp the syntactic structure.
An important but curious result is, however, that identical gains are obtained
by using trivial ``linear trees'' instead of true dependencies. The reason for
the gain thus may not be coming from the added linguistic knowledge but from
some simpler regularizing effect we induced on self-attention matrices.

\end{abstract}







\section{Introduction}

Neural machine translation (NMT) has dominated the field of MT and many works are emerging
that document that the quality of NMT can be, under some circumstances, further
improved by incorporating linguistic information from the source and/or target side.

Experiments so far were however limited to the recurrent sequence-to-sequence
architectures
\parcite{cho-EtAl:2014:EMNLP2014,bahdanau:etal:attention:iclr:2015}. \hideXXX{update, cite WMT2018 paper and Martin Popel's paper (and possibly Transformer Training Tips)} 

The recent WMT evaluations \perscite{wmt2018,findings:2019:WMT} and \perscite{CUNITransformer} show that the novel Transformer architecture
\parcite{DBLP:conf/nips/VaswaniSPUJGKP17} has set the new benchmark and it is
thus interesting to see if providing this architecture with linguistic
information is equally helpful
or if Transformer already models the phenomena
unsupervised.

We experiment with German-to-Czech and Czech-to-English translation and focus on
source-side dependency annotation using multi-task techniques.
We try two ways of forcing the model to consider source syntax: (1) by linearizing
the syntactic tree and mixing the translation and parsing training examples, and
(2) by adding a secondary objective to interpret one of the attention heads as
the syntactic tree.

In \Sref{related}, we survey recent experiments with incorporating linguistic
information into NMT, focusing particularly on works which use multi-task
learning strategies and on works which consider the syntactic analysis of the
sentence.
A brief description of the data and common settings of our experiments is
provided in \Sref{data-and-common}. In \Sref{simple-alternating}, we explore the
simple technique of multi-task by alternating training examples of the
individual tasks, discussing also the ``cost'' of multi-tasking in terms of
training steps.
\Sref{self-attention} presents the other approach, interpreting the
self-attention matrix in the
Transformer architecture as the dependency tree of the source sentence. Here we
also add the contrastive experiment with dummy diagonal parses.
\Sref{discussion} discusses the observations and we conclude in \Sref{conclusion}.

\section{Related Work}
\label{related}

The idea of multi-task training 
is to benefit from inherent and implicit similarities between two or more machine learning tasks. If the tasks are solved by a joint model with fewer or more parameters shared among the tasks, the model should exploit the commonalities and perform better in one or more tasks. This improvement can come from various sources, including the additional (often different) training data used in the additional tasks or some form of regularization or generalization that the other tasks promote.

In machine translation, multitasking has brought interesting results in
multi-lingual MT systems and also in using additional linguistic
annotation
\parcite{luong:etal:multitask:2015:arxiv,zoph-EtAl:2016:EMNLP2016,kit:multiling:iwslt:2016,johnson:zeroshot:2016}.
Similarly, \parcite{strubell-etal-2018-linguistically} incorporate
linguistic annotation to semantic role labeling task.


\citet{DBLP:conf/acl/EriguchiTC17} combined translation and dependency parsing by sharing the translation encoder hidden states with the buffer hidden states in a shift-reduce parsing model \cite{DBLP:conf/naacl/DyerKBS16}.
Aiming at the same goal, \citet{DBLP:conf/acl/AharoniG17a} proposed a very
simple method. Instead of modifying the model structure, they represented the
target sentence as a linearized lexicalized constituency tree. Subsequently, a
sequence-to-sequence (seq2seq) model \cite{DBLP:conf/nips/SutskeverVL14} was
used to translate the source sentence to this linearized tree, i.e. indeed
performing the two tasks: producing the string of the target sentence jointly
with its syntactic analysis. \citet{DBLP:conf/ijcnlp/LeMYM17} use the same trick
for (target-side) dependency trees, proposing a tree-traversal algorithm to linearize the dependency tree. Unfortunately, their algorithm was limited to projective trees.

In parallel to our work, \perscite{kiperwasser:scheduled-multitask:tacl:2018}
examined various scheduling strategies for a very simple approach to
multi-tasking: representing all the tasks converted to a common format of
source and target sequences of symbols from a joint vocabulary and training one
sequence-to-sequence system on the mix of training examples from the different
tasks. The scheduling strategy specified the proportion of the tasks in
training batches in time. \perscite{kiperwasser:scheduled-multitask:tacl:2018}
report improvements in BLEU score \parcite{BLEU} in
small-data setting for German-to-English translation for all multi-task setups
(translation combined with POS tagging and/or
source-side\footnote{\perscite{kiperwasser:scheduled-multitask:tacl:2018} do
not explicitly state whether they use the source or the target language
treebank as the training data for the parsing task. While both is actually
possible, and while even the combination of both could be tried, we assume they
used the source-side treebank only.} parsing). In the ``standard'' data size and
the opposite translation direction,
results are mixed and only one of the scheduling strategies and only the POS
secondary task help to improve MT over the baseline.

The papers mentioned so far targeted primarily the quality of MT (as measured by BLEU), not the secondary tasks. \perscite{kiperwasser:scheduled-multitask:tacl:2018} note that their system performs reasonably well in both tagging and parsing. \perscite{DBLP:conf/emnlp/ShiPK16} present
an in-depth analysis of the syntactic knowledge learned by the recurrent sequence-to-sequence NMT.
\perscite{DBLP:journals/corr/abs-1803-03585} are the first to use Transformer and observe that the recurrence is indeed
important to model hierarchical structures.

\perscite{nadejde-EtAl:2017:WMT} benefit from CCG tags
\parcite{Steedman:2000:SP} added to NMT on the source side in the form of word
factors and on the target side by interleaving the CCG tags and target words.
The additional information proves useful when the CCG tags and words are
processed in sync. \perscite{tamchyna-wellerdimarco-fraser:2017:WMT} report
similar success in interleaving words and morphological tags.


%

\section{Data and Common Settings}
\label{data-and-common}

Experiments in this paper are based on two language pairs: German-to-Czech
(de2cs) and Czech-to-English (cs2en).

German-to-Czech (de2cs)
translation is trained on Europarl \parcite{europarl} and OpenSubtitles2016
\parcite{OPUS}.
These are the only publicly
available parallel data for this language pair.
We carry out some necessary
cleanup preprocessing, character normalization
and tokenization.

Czech-to-English (cs2en) translation is
trained on a subset of CzEng 1.7
\parcite{czeng16:2016}.\footnote{\url{http://ufal.mff.cuni.cz/czeng}}
The data sizes used for MT training are summarized in \Tref{tab:data}.

For training of parsing tasks, we used the same datasets automatically
annotated on source sides. For German source we used UDPipe \parcite{udpipe},
with the model trained on Universal Dependencies 2.0 (UD, \inparcite{UD20}).
For Czech
source we used the annotation provided in CzEng release, originally created 
by Treex \parcite{tectomt:popel:2010}. This annotation is based on Prague Dependency Treebank
(PDT, \inparcite{pdt20:2006}). For parsing evaluation, we used gold test set from UD
and PDT, respectively.

\begin{table}
\begin{center}
\caption{Data used in our experiments. Test and dev data for de2cs originate
in WMT newstests. For cs2en, we use only a small portion of CzEng, as indicated
by the section numbers (\#$\cdot\cdot$).
}
\label{tab:data}
\small
\begin{tabular}{lr@{~~}r}
Dataset	& de2cs & cs2en \\
\hline
Train sent. pairs     	& 8.8M      	& \#00-\#08: 5.2M \\
Train tokens (src/tgt)	& 89M/78M  	& 61M/69M \\
Test sent. pairs         	& \llap{news 2013:} 3k      	& \#09: 10k \\ 
Dev sent. pairs          	& \llap{news 2011:} 3k      	& \#09: 1k \\
\end{tabular}
\end{center}
\end{table}



We use several automatic evaluation metrics to assess translation quality: BLEU \cite{BLEU},
CharacTER \citep{characTER},
BEER \citep{beer}, and
chrF3 \citep{chrf3}. For experiments in \Sref{simple-alternating}, the BLEU
score is cased, implemented within T2T,\furl{https://github.com/tensorflow/tensor2tensor/blob/master/tensor2tensor/utils/bleu_hook.py}
in \cref{self-attention} with
sacrebleu.\furl{https://github.com/awslabs/sockeye/tree/master/contrib/sacrebleu}
For dependency parsing task, we use 
unlabeled attachment score
(UAS).

To assess the significance of the improvement over a given baseline, we use MT-ComparEval \citep{klejch2015mt}, which implemented the paired bootstrap resampling test (confidence level 0.05 or 0.01; \inparcite{bootstrap-koehn:2004}).

\section{Simple Alternating Multi-Task}
\label{simple-alternating}




For simple alternating multi-task learning, the input and output of each task
are represented as sequences of tokens and the training examples (pairs of
sequences) are alternated in the training data.
The basic architecture for MT can thus be used
without any modifications.


\begin{figure}
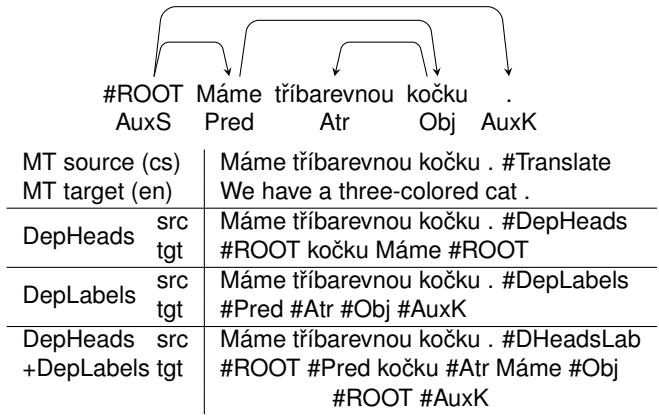

\small
\centering
        \begin{dependency}[text only label]
            \begin{deptext}
            \#ROOT \& M{\' a}me \& t{\v r}{\' i}barevnou \& ko{\v c}ku \& .  \\
            AuxS \& Pred \& Atr \& Obj \& AuxK  \\
            \end{deptext}
            \depedge[edge unit distance = 1.4ex]{1}{5}{}
            \depedge{1}{2}{}
            \depedge[edge unit distance = 2.25ex]{2}{4}{}
            \depedge{4}{3}{}
        \end{dependency}
\hspace*{-1em}
\begin{tabular}{l@{ }l|l}
\multicolumn{2}{l|}{MT source (cs)} & M{\' a}me t{\v r}{\' i}barevnou ko{\v c}ku . \#Translate \\
\multicolumn{2}{l|}{MT target (en)} & We have a three-colored cat . \\
\hline
\multirow{2}{*}{DepHeads} &src  &  M{\' a}me t{\v r}{\' i}barevnou ko{\v c}ku . \#DepHeads \\
             &tgt  &  \#ROOT ko{\v c}ku M{\' a}me \#ROOT \\
\hline
\multirow{2}{*}{DepLabels}&src  &  M{\' a}me t{\v r}{\' i}barevnou ko{\v c}ku . \#DepLabels \\
             &tgt  &  \#Pred \#Atr \#Obj  \#AuxK \\
\hline
DepHeads&src  &  M{\' a}me t{\v r}{\' i}barevnou ko{\v c}ku . \#DHeadsLab \\
+DepLabels&tgt  & \#ROOT \#Pred ko{\v c}ku \#Atr M{\' a}me \#Obj \\
&& ~~~~~~~~~~~~~~~~ \#ROOT \#AuxK \\
\end{tabular}
\caption{Sample dependency tree, inputs and expected outputs of linguistic secondary tasks.
}
\label{ling-tasks-illustr}
\end{figure}

\def\delim{~}%
\def\word{W}
\def\subw{S}
\begin{figure}
\centering
\small
\begin{tabular}{l|l}
Source words & We have a three-colored cat . \\
\hline
CountSrcWords & 6 \\
EnumSrcWords & \word{} \word{} \word{} \word{} \word{} \word{} \\
CopySrc & We have a three-colored cat . \\
\end{tabular}
\caption{Sample inputs and expected outputs of dummy secondary tasks.}
\label{dummy-examples}
\end{figure}


\subsection{Approach}

The main idea of simple alternating multi-tasking is to represent both tasks
in a formally identical way. To indicate which of the tasks should be performed
on a given input sequence, we add a special token 
%
as the very last
symbol of the sentence.

The encoder and decoder of the NMT model are thus shared for all tasks which
enables the encoder to learn source language better, but on the other hand it
occupies a certain part of the model with task alternation and
multiple language models for each task in the decoder.


\begin{figure*}
\includegraphics[width=.5\linewidth]{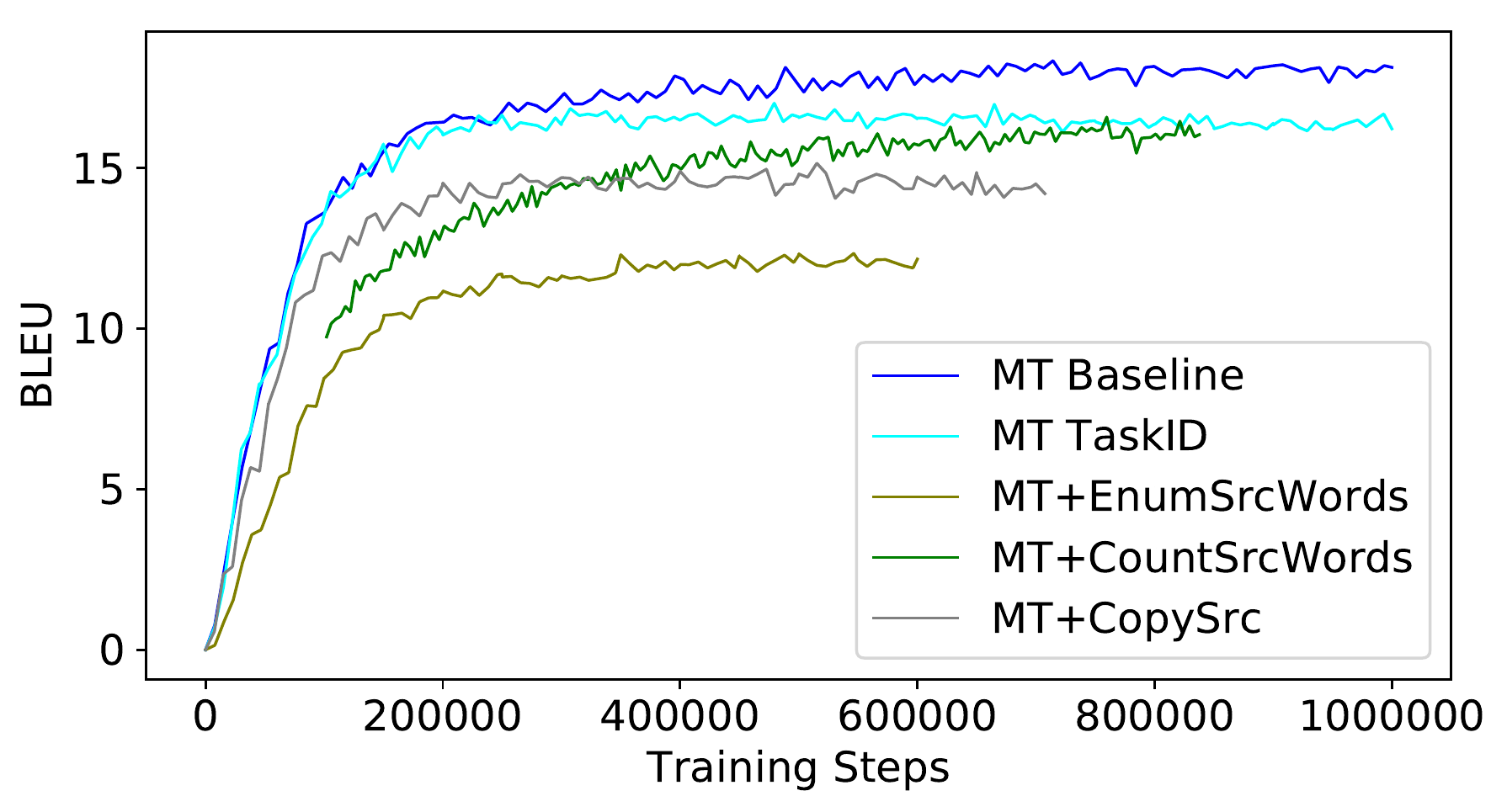}
\includegraphics[width=.5\linewidth]{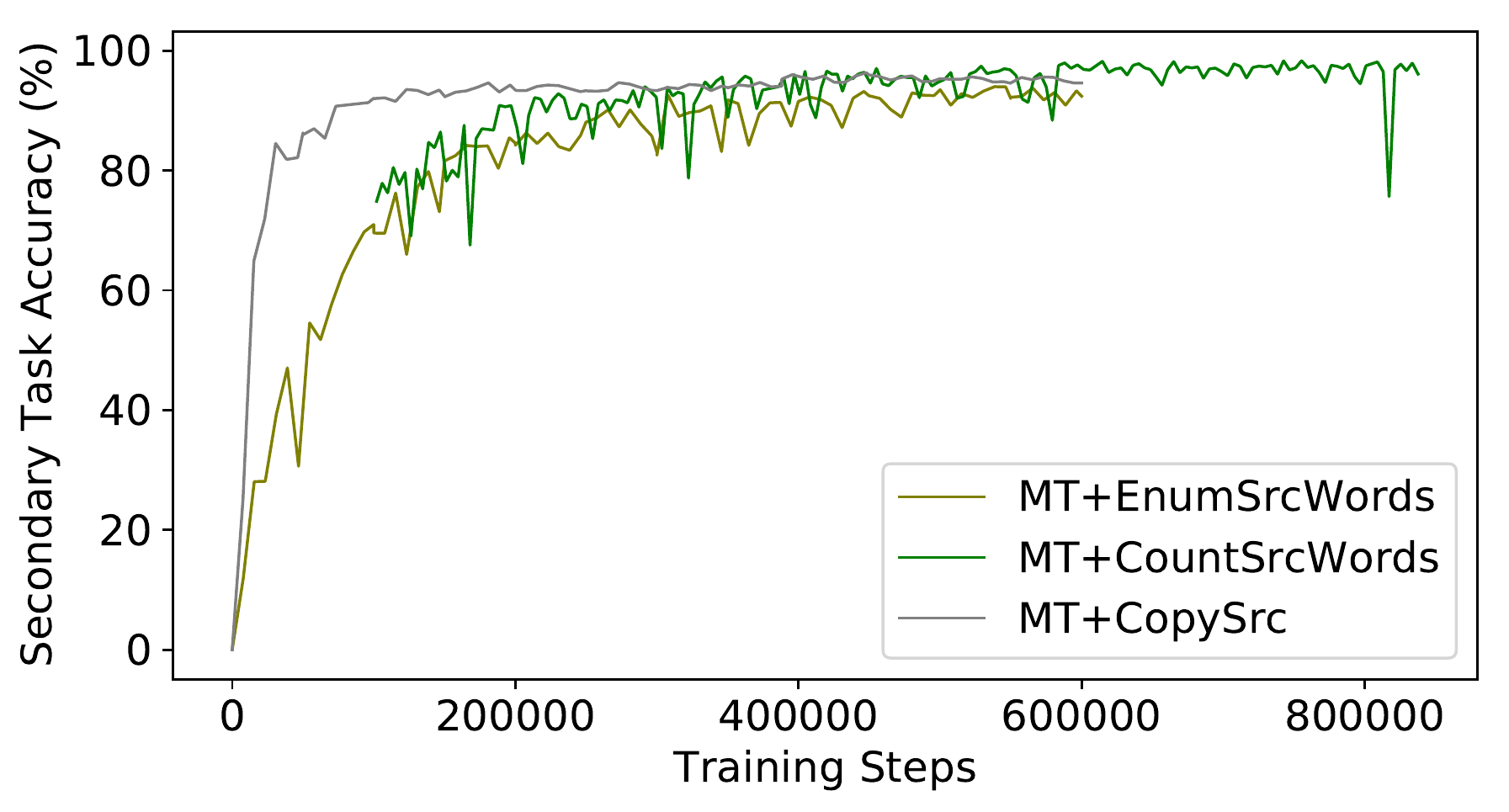}
\caption{Learning curves of the de2cs baseline and dummy secondary
tasks over training steps. MT BLEU left, percentage of correct answers
for the secondary task right.}
\label{dummy-results}
\end{figure*}

In our experiments, we mix two tasks: MT and one additional
linguistic (see \cref{ling-tasks-illustr}) or dummy referential task (see \cref{dummy-examples}). 
In \dummy{DepHeads} task, word forms of nodes' parents in the dependency tree are
predicted. We can reconstruct unlabeled dependency tree in a post-processing
step.\footnote{If one word form appears multiple times in a sentence, we attach the edge to
the nearest option. We propose this approach mostly for annotation schemes, in which
content words (in contrast to function words)  appear as inner nodes
of dependency trees. Since content words are usually not repeated in 
sentences, there is a low chance they will be mismatched.}
\dummy{DepLabels} task is tagging with dependency labels, and
\dummy{DepHeads+DepLabels} is an interleaved combination of the two.

The training data in multitasking are
selected by constant scheduler as in
\perscite{kiperwasser:scheduled-multitask:tacl:2018}, with parameter 0.5,
which means the trainer alternates between the tasks, in average, after
every training step.
As \perscite{kiperwasser:scheduled-multitask:tacl:2018} reminds, this is
different from \perscite{luong:etal:multitask:2015:arxiv} and
\perscite{zoph-knight:etal:2016} where the mixing happens at the level of
batches and not individual examples.

The experiments here in \Sref{simple-alternating} used Google's Tensor2Tensor
version 1.2.9, \texttt{transformer\_big\_single\_gpu} hyperparameter set
(hidden size 1024, filter size 4096, 16 self-attention heads, 6 layers) with
batch size 1500, 60k warmup steps and 100k shared vocabulary provided by T2T's default
SubwordTextEncoder. We note that in some cases, the particular variant of
subword units and token pre-processing can have a tremendous effect on the
final NMT performance \parcite{machacek:etal:tsd:2018}. Adding such a study is however
beyond the scope of this work.

\subsection{Training Cost of the Multi-Task}

Adding training examples of the secondary task is bound to affect the
training throughput and speed.\footnote{\label{terminology}We adopt the terminology of
\perscite{TrainingTipsfortheTransformerModel}.} The hope is that this extra
training cost is worth the gains obtained in the main task.
We examine it empirically 
by comparing the training speed of the
baseline run (no multi-task) and several versions of ``dummy'' multi-task
setups as illustrated in \Fref{dummy-examples}. In \dummy{CountSrcWords}, the
system is expected to count the source words
and emit the result
as one token holding the decimal number. \dummy{EnumSrcWords} is similar but the
expected output is much easier
for the architecture to grasp: the count should be expressed by an
appropriate number of copies of the same special output token. In
\dummy{CopySrc}, the system should simply learn to copy the source,
which should be very easy for an attentive architecture.  

The task
identification is clearly marked on input with a special token. To measure
its impact on MT quality, we provide an experimental run \dummy{MT
TaskID}, where only one MT task with task identification token is
provided.

\cref{dummy-results} summarizes the resulting learning curves on the development
set.
As we supposed, the secondary dummy task was easy to learn but
it hurts MT performance.
Enumerating and counting full words are very similar tasks in
difficulty, the model learned them almost in same time (accuracy of 80\% reached
in about 200k training steps), but enumerating
worsens MT quality much more (BLEU scores around 12 instead of beyond 15). It probably employs bigger part of decoder.
A surprising result is that the task identification token
on baseline MT data decreases overall MT performance in the long run, see the
curve ``MT TaskID'' in \cref{dummy-results}.

\begin{figure*}
\includegraphics[width=0.5\linewidth]{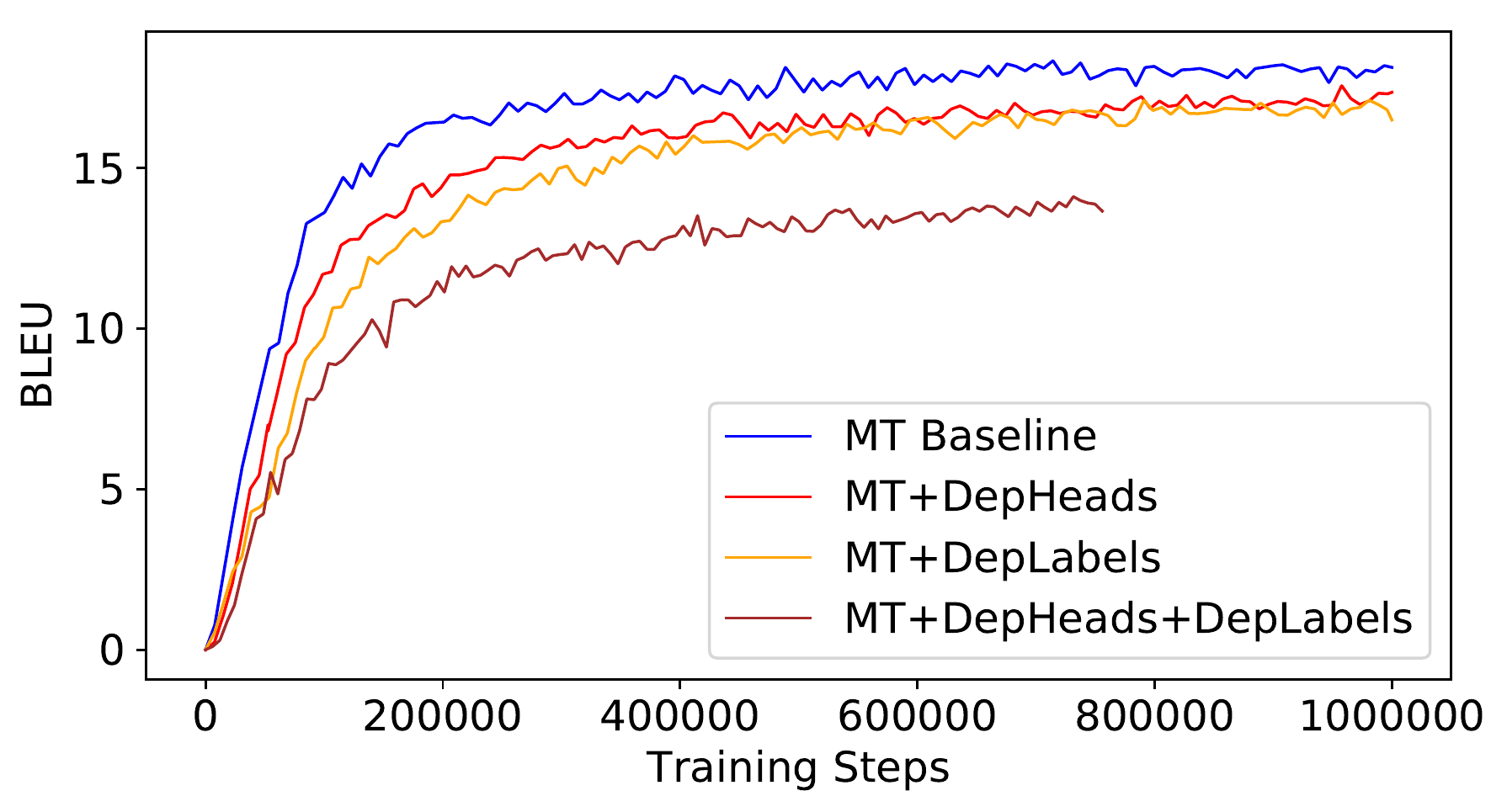}
\includegraphics[width=0.5\linewidth]{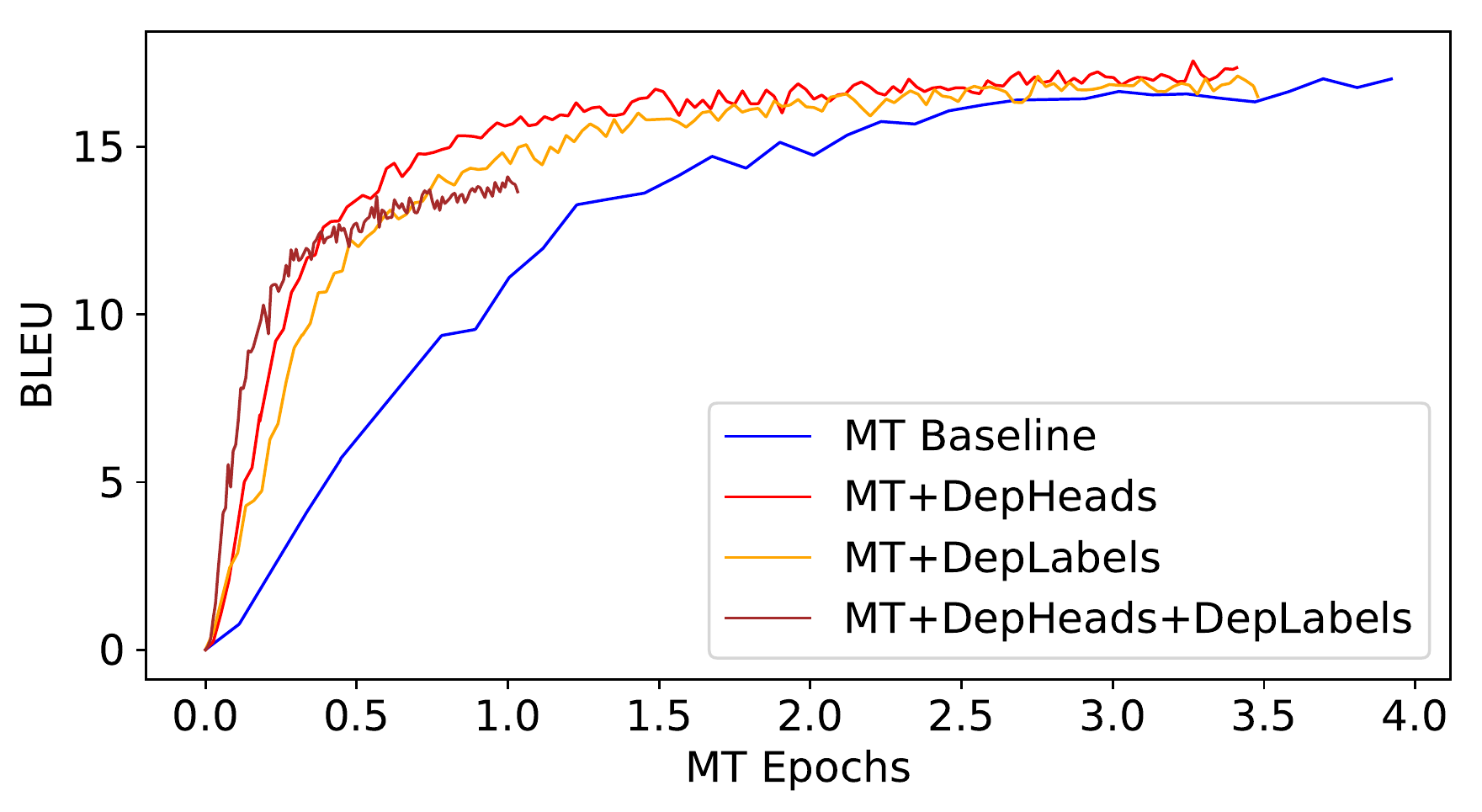}
\caption{Learning MT BLEU curves of the de2cs baseline and linguistic secondary
tasks over training steps (left) and over MT epochs (right). \hideXXX{for
the top graph, the x-axis corresponds closely to the hours of training,
doesn't it? Is the correspondence very similar for all the tasks, i.e. that
600k trainings steps was say 20h for all the setups? If yes, let's add
a secondary x axis with training hours.}} \label{fig:best-results}
\end{figure*}

\def\plusminus{$\pm$}
\def\best#1{\textbf{#1}}
\def\second#1{\textsl{#1}}
\def\mctwo#1{\multicolumn{2}{c}{\bf #1}}
\begin{table*}
\caption{Automatic scores for MT with multi-task by simple alternation. All
experiments are after 600k of training steps.
Scores (space-delimited within each cell): BLEU, CharacTER, BEER, and chrF3.
Best in bold, second-best slanted. 
Statistical significance marked as $\dag$ ($p < 0.05$) and $\ddag$ ($p < 0.01$) when compared to the second-best.
}
\label{alter-results-ling}
\small
\hspace*{-1.5em}
\begin{tabular}{l@{~~}c@{~~}c|c@{~~}c}
~             &  \mctwo{\bf~de2cs}                     &  \mctwo{\bf~cs2en} \\
\bf~Model     &  \bf~dev                            &  \bf~test        &  \bf~dev           &   \bf~test  \\
\hline
MT~Baseline   &  \best{17.90$\ddag$~60.73~52.30~47.06}             &  \best{19.74$\ddag$     58.60           53.08           48.62}&         \best{44.92~42.11~63.32~65.34}  &                                                            \best{44.20  41.68           62.70  63.88}          \\
MT+DepLabels  &  \second{16.52}~62.67~\second{50.86}~45.18  &  \second{17.87}  \second{59.65}  \second{51.67}  \second{47.01}  &                               \second{41.98}~\second{43.28}~\second{61.80}~\second{63.63}  &            \second{41.94}  42.54  \second{61.63}  \second{61.96}  \\
MT+DepHeads   &  16.36~\second{62.55}~50.76~\second{45.21}  &  17.51           62.15           51.29           46.52&          40.72~43.75~61.85~62.78         &                                                            41.10        \second{42.30}  61.41  61.68\\
MT+DepHeads+  &  13.62~70.25~48.52~43.06                    &  15.45           67.14           49.69           44.79&          39.57~45.63~60.50~61.30         &                                                            40.25        43.63           60.97  61.05           \\
~~~~~DepLabels  &  \\                                           
\end{tabular}
\end{table*}

\subsection{Results of Simple Alternating Multi-Task}


As \Tref{alter-results-ling} and \Fref{fig:best-results} (left) indicate, none of simple
alternating multi-tasking method with linguistic secondary task outperformed
baseline MT on any of our language pairs after the same amount of time. 
(In our conditions, training steps and training time are easily convertible;
600k training steps correspond to
approximately 40 hours.)

However, if we measure the performance on MT training data throughput (the
amount of data consumed in training,
\parcite{TrainingTipsfortheTransformerModel}), we see that
the multi-tasking runs achieved the same level as the baseline with less training
data.
We conclude that the cost for sharing encoder and decoder between two tasks is higher
than benefits from additional linguistic resources, but in particularly small data
settings multi-tasking may be desirable.

\cref{alt-results} shows the comparison between linguistic and dummy
secondary tasks. \dummy{DepHeads} and \dummy{DepLabels} outperformed
\dummy{CountSrcWords} and all other dummy tasks. This leads us to the conclusion
that syntactic information \emph{is} useful for the model, but the cost of the
secondary task cancels all the gains. The yet a little worse result for
\dummy{MT+DepHeads+DepLabels} confirms this: two additional tasks are more
expensive than one.

\begin{table}
\caption{Comparison of BLEU scores at 600k training steps for linguistic and
dummy secondary tasks with simple alternating approach.
}
\label{alt-results}
\small
\centering
\begin{tabular}{l|rr}
\bf~Model & \bf~dev & \bf~test \\
\hline
MT~Baseline            &  \best{17.90}  & \best{19.74} \\
MT~TaskID              &  16.53  & 18.20 \\
MT+DepLabels           &  16.52  & 17.87 \\
MT+DepHeads           &  16.36  & 17.62 \\
MT+CountSrcWords       &  15.70  & 17.51 \\
MT+CopySrc           &  14.73  & 16.07 \\
MT+DepHeads+DepLabels  &  13.62  & 15.45 \\
MT+EnumSrcWords        &  12.16  & 14.04 \\
\end{tabular}
\end{table}

\cref{alter-parsing-results} shows the performance in parsing in terms of
unlabelled attachment score and/or label accuracy, depending on the
available output
of the secondary task(s).
As the referential parser, we use UDPipe for German, the one which supervised our model.
It should be noted that we used
the supervision by UDPipe in a non-standard way. Our system (and the referential
parser) take raw word tokens on input, while
UDPipe is designed to segment multi-word tokens, such as the German \textit{zum}, into
syntactic words, as \textit{zu dem}, each of which are single nodes in tree.
For Czech, we report the score of winner in
CoNLL Shared Task 2007 \parcite{connl2007}, the latest available evaluation 
on same data. We expect that the state of the art is higher nowadays. The
limitation of our model may be the shared decoder and potentially inaccurate
automatically annotated training data.

Our systems reach a very similar UAS performance in de2cs (62.87 vs. 62.15) and
somewhat worse in cs2en (86.28 vs. 80.35) and they even surpass the referential
parser in label accuracy. The three-task system \dummy{MT+DepHeads+DepLabels} generally performs worse not
only in MT as discussed above but also in the secondary tasks.

\begin{table}[t]
\caption{Test set scores for parsing source language (German and Czech, resp.) by simple alternation. ``label acc'' is the accuracy
of tagging words with their dependency labels. Best in bold, second-best slanted.
}
\label{alter-parsing-results}
\small
\begin{tabular}{l|r@{~~}r@{~~}r@{~~}r@{~~}rrr}
~                &  \multicolumn{2}{c}{\bf~de2cs} ~  &  \multicolumn{2}{c}{\bf~cs2en} ~       \\
\bf~Model        &  \bf~UAS   &        \bf~label~acc   &  \bf~UAS  &        \bf~label~acc \\
\hline
referential~parser  &  \best{62.87}    &  \second{73.62}  &  \best{86.28}  &  83.38         \\
MT+DepLabels        &  --              &  \best{75.40}    &  --            &  \best{85.01}  \\
MT+DepHeads         &  \second{62.15}  &  --              &  \second{80.35}         &  --            \\
MT+DepHeads+        &  54.98           &  68.44           &  80.01         &  \second{83.99}         \\
~~~~~DepLabels   &  \\
\end{tabular}
\end{table}

\section{Promoting Dependency Interpretation of Self-Attention}
\label{self-attention}


In this section, we propose a different but similarly simple technique to
promote explicit knowledge of source syntax in the model. Our inspiration
comes from
the neural model for dependency parsing by
\perscite{DBLP:journals/corr/DozatM16}. The model produces a matrix $S(u, v)$
expressing the probability that the word $u$ is the head of $v$. The
construction of this matrix is very similar to the matrix of self-attention weights $\alpha$ in the Transformer
model. From this similarity, we speculate that the self-attentive architecture
of Transformer NMT
has the capacity to learn dependency parsing and we only need to promote a
little the particular linguistic dependencies captured in a treebank.


\subsection{Model Architecture}
\label{self-arch}

\begin{figure}[t]
\centering
 	\includegraphics[width=0.95\columnwidth]{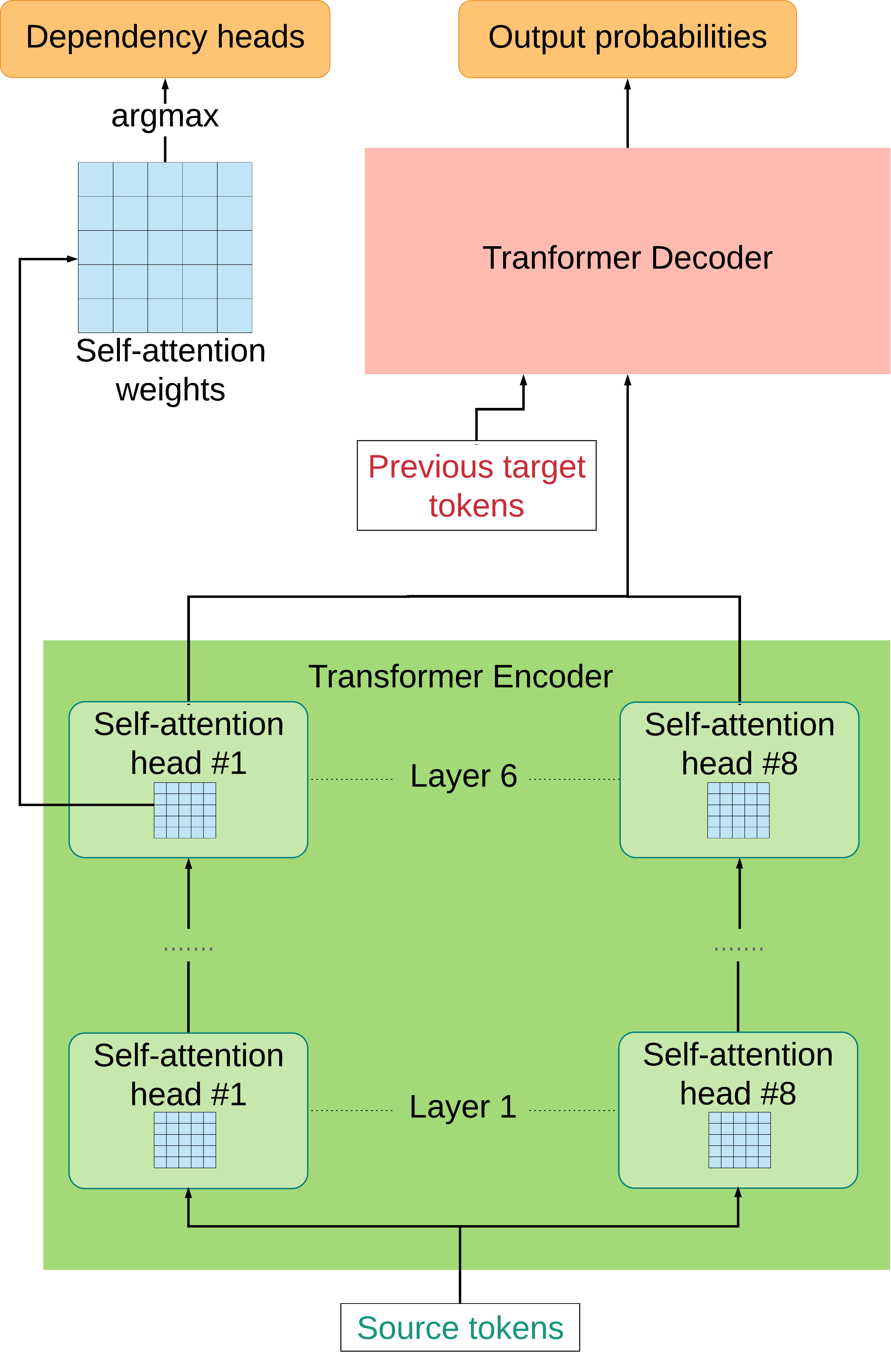}
    \caption{Joint dependency parsing and translation model (\dummy{\DepParse}).}
    \label{fig:joint_trans_depparse}
\end{figure}

\cref{fig:joint_trans_depparse} illustrates our joint model (\dummy{\DepParse}). The translation
part is kept unchanged. The only difference is that we reinterpret one of the
self-attention heads in the Transformer encoder as if it was
the dependency matrix $S(u, v)$. The training objective is combined and
maximizes (1) the translation quality in terms of cross-entropy of the
candidate translation against the reference and (2) the unlabeled attachment score (UAS) of the proposed
head against the (automatic) golden parse.
%
The particular choice of the Transformer head which will serve as the dependency parser is
arbitrary. Put differently, we constrain the Transformer model to use one of its
heads to follow the given syntactic structure of the sentence.

It would be also
possible to use e.g. the deep-syntactic parse of the sentence (the
tectogrammatical layer as defined e.g. for the Prague Dependency Treebank,
\inparcite{pdt20:2006}); we leave that for future work.

\subsection{Experiment Setup}
\label{self-setup}

Experiments in this section were carried out with T2T version 1.5.6 at the
\emph{word level}, i.e. without using subword units. We decided for this
simplification for an easier alignment between the translation and parsing
tasks.

The Transformer hyper-parameter set
\texttt{transformer\_base} \cite{TrainingTipsfortheTransformerModel} was used
for all model variants with hidden size 512, filter size 2048, 8 self-attention
heads and 6 layers in each of the encoder and decoder. From now on, we
refer the Transformer model with this hyper-parameter set as
\dummy{\transformerbase}, our baseline.
We also experimented with the choice of the encoder layer, which we use for parsing. 

In addition to the standard preprocessing for MT, we inserted a special
``ROOT'' word to the beginning of every sentence, so that the selected
self-attention head would be able to represent a dependency tree correctly.

\subsection{Layer Choice}
\label{self-layer}

\begin{table}
  \centering
    \caption{\DepParse's results in translation (BLEU) and parsing (UAS) on
  automatically annotated (\cs2en). All test BLEU gains, except for layer 0, are statistically significant with $p<0.01$ when compared to \transformerbase.
  }
  \label{tab:trans_depparse_res}
  \small
  \begin{tabular}{lcc|cc}
    &  \multicolumn{2}{c|}{\bf BLEU} & \multicolumn{2}{|c}{\bf UAS} \\
    & \bf Dev & \bf Test & \bf Dev & \bf Test \\
    \hline
    \transformerbase & 37.28 & 36.66 & -- & -- \\
    \hline
    Parse from layer 0 & 36.95 & 36.60 & 81.39 & 82.85 \\
    Parse from layer 1 & \bf 38.51 & \bf 38.01 & 90.17 & 90.78 \\
    Parse from layer 2 & 38.50 & 37.87 & 91.31 & 91.18 \\
    Parse from layer 3 & 38.37 & 37.67 & 91.43 & 91.43 \\
    Parse from layer 4 & 37.86 & 37.60 & \bf 91.65 & \bf 91.56 \\
    Parse from layer 5 & 37.63 & 37.67 & 91.44 & 91.46 \\
  \end{tabular}
\end{table}

Firstly, we experiment with selection of one of the six encoder layers from which we take
the self-attention head that will
serve as the dependency parse.
\cref{tab:trans_depparse_res} presents the results for both translation and
parsing.

It is apparent that layer 0 (the first layer) is a too early
stage for both tasks. The self-attention mechanism has only access to input word
embeddings, and their relations are very likely to be useful semantically rather
than syntactically. On the other hand, layers
1 and 2 perform well in parsing, and they are the best layers for translation
quality. A possible explanation is that they already have sufficient information
for a reasonably precise parse and do not consume the encoder's capacity for
translation. Further layers perform generally better and better in parsing
(because they are more informed) and maintain a solid performance in
translation, but the translation quality is slowly decreasing.
For the following, we select layer 1 to demand syntactic information from.

\subsection{Performance in Translation}
\label{self-mt}

\begin{table}[t]
\caption{BLEU scores on test set for translation task (T2T 1.5.6, word level). Statistical significance marked as $\dag$ ($p < 0.05$) and $\ddag$ ($p < 0.01$) when compared to \transformerbase.}
\label{multidec-results}
\begin{center}
\small
\begin{tabular}{lcc}
\bf Model        	& \bf de2cs	& \bf cs2en	\\
\hline
\transformerbase    & \second{13.96}	&  \second{36.66} \\
Alternating multi-tasking (\cref{simple-alternating})  		& 12.85 &  36.47\\	
\DepParse{} (\cref{self-attention})		& \best{14.27}$^\dag$	&  \best{38.01}$^\ddag$ \\
\end{tabular}
\end{center}
\end{table}

\Tref{multidec-results} compares the performance of the baseline Transformer,
the simple alternating setup from \Sref{simple-alternating} (DepHeads src) and the multi-task
setup from this section. All these runs use T2T version 1.5.6 and use
words, not subword units. This also explains the decrease in BLEU compared to
\Tref{alter-results-ling}. 
The \DepParse model significantly outperforms the baseline (38.01 vs. 36.66 and 14.27 vs. 13.96).

\subsection{Performance in Parsing}
\label{self-parsing}

\begin{table}
\caption{UAS on gold annotated test sets for parsing task.}
\label{multidec-results-parse}
\begin{center}
\small
\begin{tabular}{lcc}
\bf Model        	& \bf de2cs	& \bf cs2en	\\
\hline
Referential parsers 	& 62.87 &  86.28 \\ 
\DepParse				& 76.48	&  82.53 \\
\end{tabular}
\end{center}
\end{table}

In addition to the automatically annotated dev and test set, we also evaluated
our model on the gold evaluation sets
from UD 2.0 for German and from PDT 2.5 for Czech. The referential parsers were
the same as for \cref{alter-parsing-results} above. \cref{multidec-results-parse} shows
that our model achieved good results in comparison to the baseline model on those datasets, even though ours was trained using synthetic data.


\subsection{Diagonal Parse}
\label{self-diag}
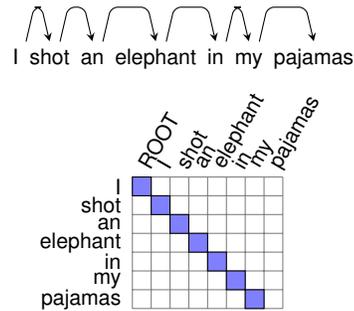
\begin{figure}[t]
\centering
\footnotesize
        \begin{dependency}[text only label]
            \begin{deptext}
            I \& shot \& an \& elephant \& in \& my \& pajamas \\
            \end{deptext}
            \depedge{1}{2}{}
            \depedge{2}{3}{}
            \depedge{3}{4}{}
            \depedge{4}{5}{}
            \depedge{5}{6}{}
            \depedge{6}{7}{}
        \end{dependency}
            \begin{tikzpicture}[scale=0.25]
                \draw[step=1cm,draw=gray] (0,0) grid (8,7);
                
                \foreach \f [count=\y] in {I, shot, an, elephant, in, my, pajamas} {
                    \node[left] at (-.2,7.5-\y) {{\raggedleft \f }};
                }
                
                \foreach \e [count=\x] in {ROOT, I, shot, an, elephant, in, my, pajamas} {
                    \node[rotate=60,right] at (\x-.6,7.2) {{\raggedright \e}};
                }
                
                \foreach \x [count=\y] in {0, 1, 2, 3, 4, 5, 6} {
                    \draw[fill=blue!50] (\x,7-\y) rectangle +(1,1);
                }
            \end{tikzpicture}
    \caption{Dummy dependencies with diagonal matrix (the columns represent the heads, the rows are dependents).}
    \label{fig:monotree-vs-matrix}
\end{figure}

For contrast, we conduct an experiment with a simpler sentence structure, which
we call the ``diagonal parse''. In the diagonal parse, the dependency head of a token
is simply the previous token, i.e. a linear tree, as illustrated in \cref{fig:monotree-vs-matrix}.

Our model for the joint diagonal parsing and translation (\dummy{\DiagonalParse}) is
identical to the \dummy{\DepParse{}} model, which has been described in  \cref{self-arch}.
We only use diagonal matrices during training, instead of the dependency matrices.
The main goal of this setup is to examine 
the benefits of the additional syntactic information for machine translation. 

\begin{table}[t]
\small
  \caption[\DiagonalParse's results in translation (BLEU) and diagonal parsing (precision) on \cs2en.]{\DiagonalParse's results in translation (BLEU) and diagonal parsing (precision) on \cs2en. All test BLEU improvements are statistically significant with $p<0.01$ when compared to the \transformerbase.}
  \label{tab:res-translate-monoparse}
\centering
\vspace{2ex}
  \begin{tabular}{lcc|cc}
    &  \multicolumn{2}{c}{\textbf{BLEU}} & \multicolumn{2}{|c}{\textbf{Precision}} \\
    & \textbf{Dev} & \textbf{Test} & \textbf{Dev} & \textbf{Test} \\
    \hline
    \transformerbase & 37.28 & 36.66 & -- & -- \\
    \hline
    Parse from layer 0 & 38.68 & \textbf{38.14} & 99.97 & 99.96 \\
    Parse from layer 1 & \best{39.11} & 38.06 & \best{99.99} & \textbf{99.99} \\
    Parse from layer 2 & 37.85 & 37.85 & 99.98 & 99.98 \\
    Parse from layer 3 & 37.93 & 37.70 & 99.97 & 99.98 \\
    Parse from layer 4 & 37.68 & 37.47 & 99.98 & 99.96 \\
    Parse from layer 5 & 37.53 & 37.54 & 99.96 & 99.95 \\
  \end{tabular}
\end{table}


\cref{tab:res-translate-monoparse} documents that the \dummy{\DiagonalParse} model is
very effective.  The diagonal parsing precision is, as expected, very high,
ranging from 99.95\% to 99.99\% on the test set.  This joint model also
outperformed the baseline in translation task with all its variants (BLEU
scores vary from 37.47 to 38.14, compared to 36.66).

Moreover, these results form an observable pattern, in which the best result
comes from the model with parsing with the head on layer 0.
Parsing from deeper layers still helps to improve translation over baseline, but the BLEU scores decrease.
We believe that a possible explanation for this pattern is that the diagonal matrix represents the relation between the preceding token and the current token.
This simple sentence structure can serve as an additional positional information to the absolute positional embeddings.
Therefore, the sooner the model is forced to recognize this positional
information (via training the parsing task), the better it can learn to translate.
Another possible explanation is the regularization effect of the
diagonal parse.

\begin{figure*}[t]
\centering
	\includegraphics[width=0.85\textwidth]{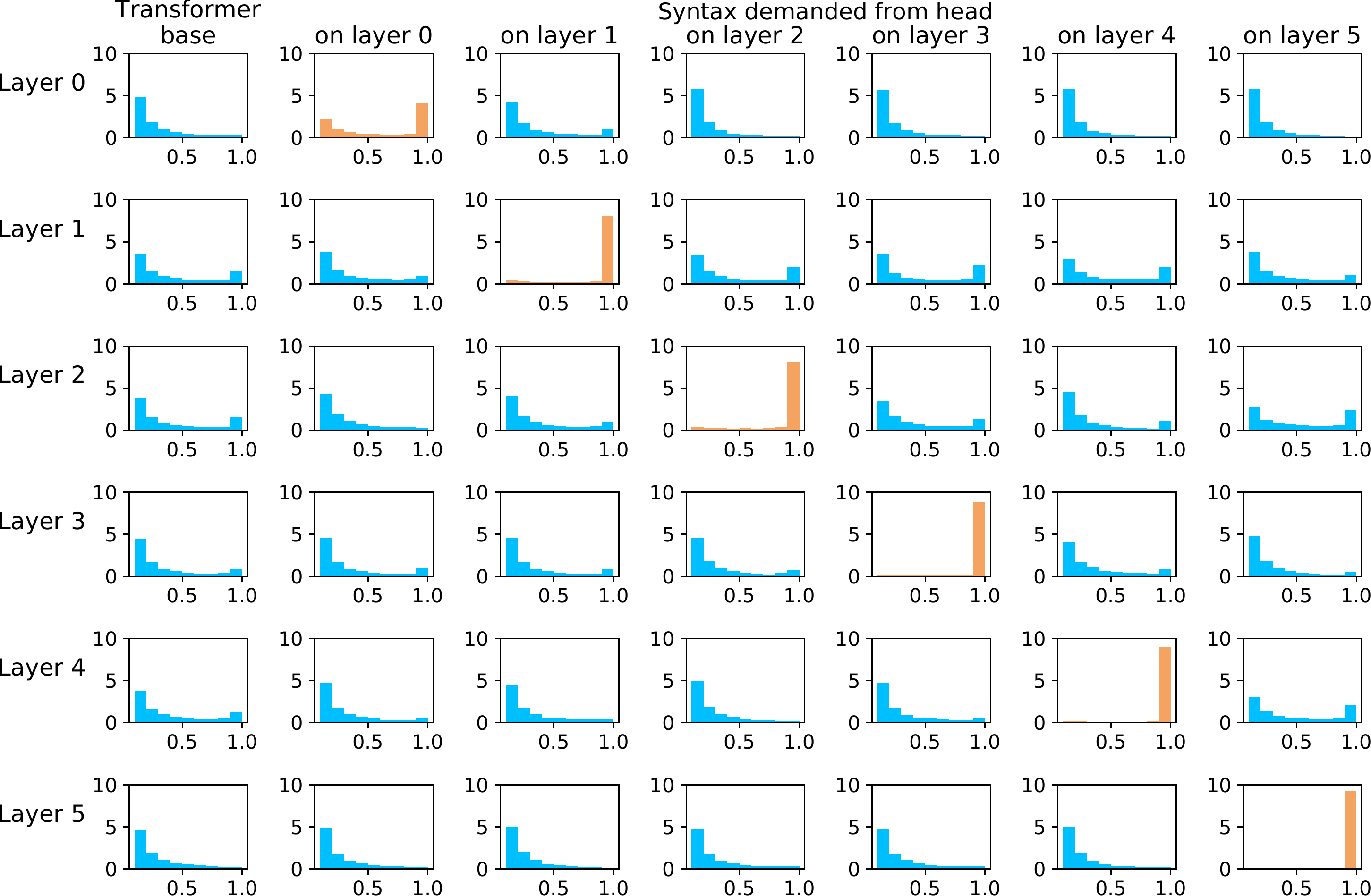}
    \caption{Histogram of normalized self-attention weights in the encoder.}
    \label{fig:att_dist}
\end{figure*}

\subsection{Training Speed}


As seen above, the secondary objective for both true dependency tree as well as
for the diagonal parse help to improve translation quality in terms of BLEU
score.

It is worth noting that this extension of the model, the secondary objective,
comes at a rather small cost in training time compared to the baseline
Transformer.


The training
time (including internal evaluation every 1000 steps)
on a single GPU NVIDIA GTX 1080 Ti
needed to reach 250k steps for \transformerbase was
about 1 day and 4 hours while our joint models needed only 10\%-13\% more time to train on both tasks.

\subsection{Self-Attention Patterns in the Encoder}

\begin{figure*}[t]
\centering
	\includegraphics[width=0.85\textwidth]{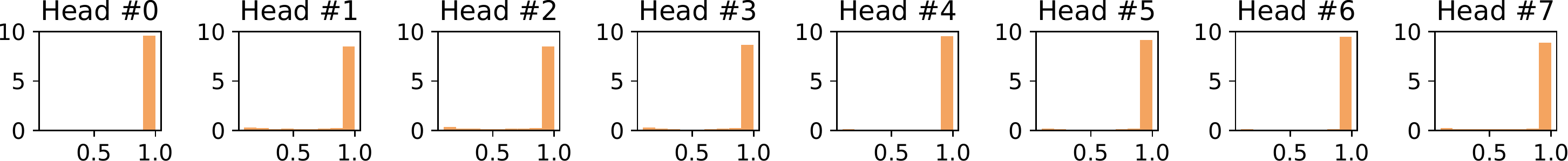}
    \caption{Histogram of self-attention weights in the encoder's layer 4 when parsing from layer 4.}
    \label{fig:att_dist_4}
\end{figure*}

\cref{fig:att_dist} presents the behavior of self-attention mechanism in
each layer of our models for the first 100 sentences in the test
set.
We summarize the self-attention across the sentences and Transformer heads of a
given layer using a histogram of
observed self-attention weights. Most cells in the attention matrix indicate no
attention, so the bin $[0.0,0.1)$ always receives the highest value in the
histogram. For clarity of the picture, we exclude this bin and focus on other
observed values of attention weights.


As can be seen from \cref{fig:att_dist}, the Transformer base encoder shows very
similar distribution of attention weights across all its six layers: most cells
have low values. 
The distributions are very different, when one of the heads at a given layer is
trained to perform source-side parsing, i.e. to assign exactly one governor
to each word in the input sentence. With this constraint, the attention
distribution \emph{at the particular layer} is peaked, with each position
attending clearly to only one or two positions from the previous layer.

This behavior is apparent in all our multi-task models except the ``Parse
from layer 0", where we see a mix of the baseline and the peaked pattern. As mentioned in \cref{self-mt}, this model performed
badly on both tasks. While the causality is unclear, we at least see that the
sharpness in attention is related to the better performance.

\cref{fig:att_dist_4} documents another interesting observation (as above, the
bin$[0.0,0.1)$ was excluded). One could perhaps
expect such a sharp attention from the one particular head which was trained to predict dependencies  but interestingly, the same sharpness is observed in all heads of the given layer.
A possible reason may be the vector concatenation and layer normalization after each multi-head attention layer in the Transformer.


\section{Discussion}
\label{discussion}



\perscite{kiperwasser:scheduled-multitask:tacl:2018} suggest another
representation of \dummy{DepHeads}, which doesn't suffer on unknown words and
repeated words. They indicate the governing node as an offset from the node's position
represented as decimal number, positive to the right, negative to
the left. As we documented in \cref{simple-alternating}, Transformer can easily
learn to count words, so this
representation should be considered in future work.

We let aside a question of vocabulary design for multi-tasking. In T2T's
SubwordTextEncoder (STE), the
vocabulary is constructed automatically from a training data sample, so that
frequent words are represented as single subwords and rare words as sequences
of characters.
We assume that balancing the importance of source and target sides of particular
tasks
in STE input, thus steering the vocabulary towards more efficient
representations of some of these sides, could lead to better quality. This could
be further combined with
various parameters for the constant task scheduler.

Multiple multi-task experiments (\inparcite{Niehues2017ExploitingLR}, \inparcite{scarce-mtl},
etc.) mention notable 
gains on small data scenarios.
As documented by \perscite{koehn-knowles:2017:NMT}, under certain training data
size, NMT is actually much worse than conventional phrase-based MT. It is
unclear if the gains from NMT multi-tasking are obtained also after this
critical corpus size, or if they are limited to the data sizes where NMT is
not so effective.

The observation in \cref{self-diag} that a very similar gain BLEU can be
achieved using either true syntactic trees or dummy diagonal parse is casting
doubt on the utility of explicit linguistic information for Transformer models.
While we keep the stance that linguistic generalization will be useful for
translation quality in the long term, our results cannot confirm this yet. We
are nevertheless happy that we included this dummy experiment and opened this
concern, rather than confidently claiming that source syntax helps Transformer.
We share the opinion with \perscite{dummies:2019} that dummy baselines are critically needed
for trustworthy progress in NMT.


\hideXXX{
 encoding is different, they use LSTM (where the task
indication may perform better than with transformer, esp. since we add the token
to the input, not to the output); they use 4M sent pairs as the "large" data
setting and 4M tokens as the "small" data setting. Both are smaller than our
data.  Another difference is that they train the parsing subtask on a different
corpus: they extract the sentences and trees from the gold-annotated treebank,
and mix them with sentence pairs from the parallel corpus. We use the same
underlying sentences (coming from a parallel corpus) and annotate them with
automatic tools.
Yet another thing: Their small data setting may benefit from the additional
source-side monolingual data, not from the POS-tagging! (Because the MT data is
so small that the POS corpus contributes significantly)
}

One limitation of our setup was that
our model was trained on automatic parses.
Hence, it would be interesting to fine-tune our model with 
gold-annotated trees, which could lead to a better parsing
performance. We leave this for future work.

\section{Conclusion}
\label{conclusion}

We proposed two techniques of promoting the knowledge of source syntax in the
Transformer model of NMT by multi-tasking and evaluated them at reasonably large
data sizes.

The simple data manipulation
technique, alternating translation and linearized parsing, is impractical. Learning to translate and parse improves over comparable multi-task setups with
uninformative (``dummy'') secondary tasks, but overall it performs worse than
single-task translation model. In low-resource conditions, the gain from the
multi-tasking may be useful.

The other technique, re-interpreting one of the self-attention heads in the
Transformer model as the dependency analysis of the sentence, is surprisingly
effective. At little or no cost in training time, Transformer learns to
translate and parse at the same time. The parse accuracy is reasonable and the
translation is significantly better than the baseline. Curiously, very similar
gains can be obtained by predicting a ``diagonal parse'', i.e. linguistically
uninformed linear tree. The full explanation of this behavior is yet to be
sought for.

\hideXXX{EMAIL:
Guys,
I think you should also consider adding Strubel et al. 2018
(Linguistically-Informed Self-Attention for Semantic Role Labeling, 
\perscite{strubell-etal-2018-linguistically}
https://arxiv.org/abs/1804.08199) to Related Work section. Since it was one
of the best papers in EMNLP 2018, the reviewers will probably know it.
Best, Jana Strakova}

\section*{Acknowledgments}


The research was partially supported by the grants
19-26934X (NEUREM3) of the Czech Science Foundation 
and H2020-ICT-2018-2-825460 (ELITR) of the EU. 


%
%
%
%
%
%
%
%
%
%
%
%
%
%
%
%
%
%
%
%

\small{
\bibliographystyle{cys}
\bibliography{biblio}
}
\normalsize

\begin{biography}[]{} 
\end{biography}

{\vskip 12pt}
\noindent
\footnotesize {\textit{
Computación y Sistemas, Vol. 23, No. 3, 2019, pp. 923–934 \\
Article received on 23/02/2019; accepted on 04/03/2019.\\
Corresponding author is Ond{\v r}ej Bojar. \\
doi: 10.13053/CyS-23-3-3265
}}

\end{document}